\documentclass[sigconf]{acmart}

\AtBeginDocument{%
  \providecommand\BibTeX{{%
    \normalfont B\kern-0.5em{\scshape i\kern-0.25em b}\kern-0.8em\TeX}}}

\renewcommand\footnotetextcopyrightpermission[1]{} 

\usepackage{subfigure}
\usepackage{multirow}
\begin{document}
\fancyhead{}

\title{Robust Fine-tuning for Pre-trained 3D Point Cloud Models}


\author{Zhibo Zhang}
\affiliation{%
  \institution{School of Computer Science, Fudan University}
  \city{Shanghai}
  \country{China}}
\email{zhibozhang21@m.fudan.edu.cn}

\author{Ximing Yang}
\affiliation{%
  \institution{School of Computer Science, Fudan University}
  \city{Shanghai}
  \country{China}}
\email{xmyang19@fudan.edu.cn}

\author{Weizhong Zhang}
\affiliation{%
  \institution{School of Data Science, Fudan University}
  \city{Shanghai}
  \country{China}}
\email{zhangweizhongzju@gmail.com}

\author{Cheng Jin}
\affiliation{%
  \institution{School of Computer Science, Fudan University}
  \city{Shanghai}
  \country{China}}
\email{jc@fudan.edu.cn}

\begin{abstract}
This paper presents a robust fine-tuning method designed for pre-trained 3D point cloud models, to enhance feature robustness in downstream fine-tuned models. We highlight the limitations of current fine-tuning methods and the challenges of learning robust models. The proposed method, named Weight-Space Ensembles for Fine-Tuning then Linear Probing (WiSE-FT-LP), integrates the original pre-training and fine-tuning models through weight space integration followed by Linear Probing. This approach significantly enhances the performance of downstream fine-tuned models under distribution shifts, improving feature robustness while maintaining high performance on the target distribution. We apply this robust fine-tuning method to mainstream 3D point cloud pre-trained models and evaluate the quality of model parameters and the degradation of downstream task performance. Experimental results demonstrate the effectiveness of WiSE-FT-LP in enhancing model robustness, effectively balancing downstream task performance and model feature robustness without altering the model structures.
\end{abstract}






\maketitle

\section{Introduction}

Over the past few years, a fundamental goal of machine learning has been to develop reliable models that apply to a wide range of data distributions. In research in this area, various distribution shifts have been proposed and current algorithmic approaches have been found to gain little in terms of enhanced robustness\cite{DBLP:conf/nips/TaoriDSCRS20, DBLP:conf/icml/MillerTRSKSLCS21}. Although these negative results highlight the difficulty of learning robust models, recent large-scale pre-trained models such as CLIP \cite{DBLP:conf/icml/RadfordKHRGASAM21}, ALIGN \cite{DBLP:conf/icml/JiaYXCPPLSLD21} and BASIC \cite{DBLP:journals/ijon/PhamDGKLYYCLWTL23}) demonstrates unprecedented robustness to these challenging distribution shifts. The success of these models suggests that pre-training on large, heterogeneous datasets is a promising direction for increasing robustness. However, an important limitation is that these robustness improvements are most significant in a Zero-/Few-Shot Learning setting, where the target distribution is not fine-tuned while the model is inferring, or only on tiny targets. Fine-tune the distribution.

In specific applications, pre-trained models can be fine-tuned on additional application-specific data, often resulting in large performance improvements on the target distribution. However, in experiments with CLIP \cite{DBLP:conf/icml/RadfordKHRGASAM21} and BASIC \cite{DBLP:journals/ijon/PhamDGKLYYCLWTL23}, fine-tuning comes at the cost of reduced robustness: among several natural distribution shifts, The accuracy of their fine-tuned model is lower than the original pre-trained model. This leads to a natural question: can a pre-trained model be fine-tuned without degrading accuracy under distribution shifts?

With the growing importance of pre-trained models in machine learning, techniques for fine-tuning them for downstream applications are crucial. Previous research has highlighted the challenge of achieving robustness while maintaining high accuracy in fine-tuned models, especially under distribution shifts. Despite various fine-tuning methods explored by researchers, none has consistently produced both robust and high-accuracy models across different distributions.

Researchers initially analyzed the impact of different fine-tuning methods (e.g., last-layer fine-tuning, end-to-end fine-tuning, and hyperparameter adjustments) on model accuracy under distribution shifts. They found that even slight variations in hyperparameters can result in significant differences in the robustness of fine-tuned models, making optimal hyperparameter selection challenging. Additionally, more aggressive fine-tuning strategies, such as using larger learning rates, may boost accuracy on the target distribution but could adversely affect performance under distribution shifts.

The Weight-Space Ensembles for Fine-Tuning (WiSE-FT) approach, proposed by Wortsman et al. \cite{DBLP:conf/cvpr/WortsmanIKLKRLH22}, effectively addresses these challenges by offering a robust zero-shot pre-training model fine-tuning method aimed at reconciling trade-offs and achieving optimal performance. This method involves two key steps: first, fine-tuning a zero-shot pre-trained model on the target distribution; second, combining the original pre-trained and fine-tuned models through a weight-space ensemble by linearly interpolating their weights. Compared to conventional methods, WiSE-FT significantly enhances accuracy under distribution shifts while maintaining strong performance on the target distribution. This approach represents a notable advancement in model fine-tuning, providing a promising solution to improve model robustness and performance across different distributions in the field.

However, existing methods like WiSE-FT have mainly focused on visual-language pre-training models with zero-shot inference, neglecting general pre-training models that necessitate training additional inference heads from scratch without zero-shot inference. This paper primarily investigates a less-explored area: point cloud pre-training models, proposing a robust fine-tuning method inspired by weight-space ensemble fine-tuning, tailored specifically for these models—Weight-Space Ensembles for Fine-Tuning then Linear Probing (WiSE-FT-LP).

WiSE-FT-LP involves three key steps:
\begin{enumerate}
    \item Firstly, integrating inference heads required for downstream tasks into the pre-trained model backbone network and fine-tuning it on the target distribution.
    \item Secondly, combining the original pre-trained model with the fine-tuned model backbone using a weight-space ensemble with linear weight interpolation.
    \item  Lastly, fixing the backbone network parameters and fine-tuning only the inference head.
\end{enumerate}

This method significantly enhances accuracy under distribution shifts while maintaining or even improving performance on the target distribution compared to standard fine-tuning. Importantly, the improved robustness achieved by WiSE-FT-LP comes without incurring any additional computational cost during fine-tuning or inference. In conclusion, WiSE-FT-LP offers a straightforward, effective, and widely applicable solution for fine-tuning general pre-trained models, requiring only a few lines of code for implementation.

Specifically, this paper applies the WiSE-FT-LP method to two representative point cloud pre-training models: ReCon \cite{DBLP:conf/icml/QiDFGZMY23}, which leverages point cloud and image multi-modal pre-training simultaneously, and Point-M2AE \cite{DBLP:conf/nips/ZhangG0FZW0022}, pre-trained using multi-scale pure point cloud data. These models have demonstrated effectiveness in point cloud object classification tasks. To further enhance their robustness for practical applications, we employed the WiSE-FT-LP method for fine-tuning and evaluated the model parameter quality before and after fine-tuning. During the evaluation, we utilized two tasks: linear support vector machine (SVM) classification and few-shot classification to comprehensively assess changes in model feature robustness. Analyzing the alterations in model parameter quality provides insights into the impact of the WiSE-FT-LP method on model robustness.

Experimental results demonstrate that the WiSE-FT-LP method maintains high performance on the target distribution while enhancing model robustness. This outcome confirms the method's capability and provides valuable insights for leveraging point cloud pre-training models. Additionally, we conducted an in-depth exploration of the changes in model performance, revealing both the applicability and limitations of the method. These discussions offer valuable inspiration for further refining the approach and guiding future research in this field.

In summary, these contributions represent key advancements in our study:

\begin{itemize}
    \item Introducing the WiSE-FT-LP method, effectively striking a balance between feature robustness and downstream task performance in 3D point cloud pre-training. This method represents a novel approach to addressing the challenge of maintaining robust features while optimizing task-specific performance.
    
    \item Proposing a method for assessing backbone network robustness, providing a systematic framework to evaluate the stability and resilience of backbone architectures used in 3D point cloud pre-training models. This contributes to a deeper understanding of model behavior and performance under various conditions.
    
    \item Conducting experiments to validate the efficacy of WiSE-FT-LP on 3D point cloud pre-training models. The experimental results demonstrate that WiSE-FT-LP achieves a desirable trade-off between feature robustness and performance on downstream tasks, highlighting its practical applicability and effectiveness in real-world scenarios.
\end{itemize}

These contributions collectively advance our understanding of enhancing model robustness and performance in the context of 3D point cloud pre-training, providing valuable insights for future research and development in this domain.

\begin{figure*}[htb]
    \includegraphics[width=\textwidth]{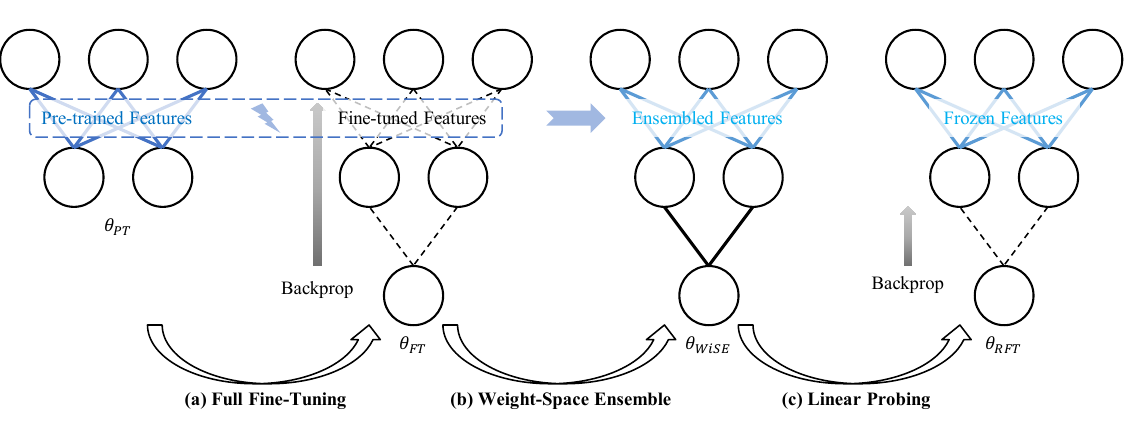}
    \caption{Fine-tuning steps for WiSE-FT-LP.}
    \label{wiseftlp}
\end{figure*}

\section{Related Works}

\subsection{Robust Fine-tuning for Pre-trained Models}

Fine-tuning a pre-trained model has become the de facto standard for transfer learning, driven by the rapid advancement of large-scale pre-trained models \cite{DBLP:conf/icml/RadfordKHRGASAM21}. Consequently, there is a growing focus on refining techniques for fine-tuning to enhance performance on downstream tasks \cite{DBLP:conf/cvpr/GeY17, DBLP:conf/cvpr/GuoSKGRF19}. However, limited attention has been given to the robustness and generalization of these fine-tuned models.

Recent studies \cite{DBLP:conf/iclr/KumarRJ0L22, DBLP:journals/tmlr/AndreassenBNR22, DBLP:conf/cvpr/WortsmanIKLKRLH22} have explored the evolution of out-of-distribution (OOD) robustness during downstream fine-tuning. They reveal that the inherent robustness of pre-trained models diminishes as they are fine-tuned for specific tasks. LP-FT \cite{DBLP:conf/iclr/KumarRJ0L22} proposes a two-step approach: first, employing linear probing to identify an optimal classification head, followed by full fine-tuning to mitigate the degradation of robust pre-trained features. WiSE-FT \cite{DBLP:conf/cvpr/WortsmanIKLKRLH22} utilizes a weight-space ensemble of original pre-trained weights and fine-tuned weights to achieve improved accuracy on both in-distribution (ID) and OOD datasets.

These advancements underscore the importance of optimizing fine-tuning for task performance while maintaining and enhancing model robustness and generalization across diverse datasets.

\subsection{Self-supervised Learning for Point Clouds}

Contrastive representation learning, a prominent self-supervised learning paradigm \cite{DBLP:conf/cvpr/HadsellCL06}, aims to learn latent semantics by leveraging constructed invariance or equivalence \cite{DBLP:conf/iclr/DangovskiJLHSCA22}. PointContrast \cite{DBLP:conf/eccv/XieGGQGL20} introduces geometric augmentation to generate positive and negative pairs, enhancing the learning of point cloud representations. CrossPoint \cite{DBLP:conf/cvpr/AfhamDDDTR22} extends this idea by incorporating both inter- and intra-modal contrastive learning techniques. Meanwhile, PointCLIP \cite{DBLP:conf/cvpr/ZhangGZLM0QG022} focuses on aligning point clouds with 2D depth images to improve image-based representation learning.

On the other hand, generative mask representation learning, as pioneered by MAE \cite{DBLP:conf/cvpr/HeCXLDG22}, requires models to reconstruct masked input data, fostering structured knowledge acquisition by associating different local patches. Point-MAE \cite{DBLP:conf/eccv/PangWTLTY22} extends this approach to point clouds, emphasizing masked reconstruction in the point cloud domain. Point-M2AE \cite{DBLP:conf/nips/ZhangG0FZW0022} enhances this further with a hierarchical Transformer architecture and refined masking strategies.

Recent innovations include ACT \cite{DBLP:conf/iclr/DongQZZSGYM23}, which employs a cross-modal autoencoder for reconstructing input data across different modalities, extracting "dark knowledge" from diverse sources. ReCon \cite{DBLP:conf/icml/QiDFGZMY23} combines cross-modal contrastive learning with masked autoencoding to enhance the model's capabilities in structured representation learning.

These pioneering approaches collectively push the boundaries of self-supervised learning, effectively tackling diverse challenges and deploying innovative strategies for robust representation learning. As the representative 3D point cloud pre-training models in this study, we selected Point-M2AE \cite{DBLP:conf/nips/ZhangG0FZW0022}, focusing solely on the 3D point cloud modality without additional modalities, and ReCon \cite{DBLP:conf/icml/QiDFGZMY23}, which integrates multiple modalities to achieve its results.

\section{Approach}

\subsection{WiSE-FT-LP}

This paper details a robust fine-tuning method called WiSE-FT-LP, which aims to improve the accuracy and robustness of deep neural networks on various tasks and datasets. The core idea of the WiSE-FT-LP method is to make full use of the knowledge of the pre-trained model and adopt a reasonable interpolation strategy in the fine-tuning process to balance the relationship between accuracy performance and robustness.

Specifically, as shown in Figure \ref{wiseftlp}, the WiSE-FT-LP method involves several key steps:

\subsubsection{Initial Setup and Fine-Tuning}
Start with a pre-trained model backbone $\theta_{PT}$, a neural network already trained on a large dataset (e.g., ShapeNet \cite{DBLP:journals/corr/ChangFGHHLSSSSX15}). Add a downstream task head (such as an MLP or task-specific layers) to $\theta_{PT}$ and initialize its parameters randomly. Fine-tune this modified model on the dataset of interest (downstream task) to obtain a fine-tuned model $\theta_{FT}$.

\subsubsection{Weight-Space Interpolation}
During the fine-tuning process, calculate an interpolation weight $\alpha$ that determines the mix between $\theta_{PT}$ and $\theta_{FT}$:
$$\theta_{WiSE}(\alpha) = (1 - \alpha)\theta_{PT} + \alpha\theta_{FT},$$
where $\alpha \in [0, 1]$. Vary $\alpha$ to generate a series of models with different interpolation weights.

\subsubsection{Freezing Backbone and Further Fine-Tuning}
Freeze the parameters of the backbone network ($\theta_{PT}$) to preserve the learned features. Further, fine-tune the downstream task head on the specific task dataset to obtain a robust model $\theta_{RFT}$. This step enhances the model's ability to generalize across different data distributions and improve robustness against noise and challenges.

\subsubsection{Selection of Optimal Interpolation Coefficient}
Evaluate the performance of each model in the series on a validation set. Choose the interpolation coefficient (value of $\alpha$) that maximizes the improvement in accuracy while maintaining model robustness. This selection balances task-specific performance gains with overall model stability and adaptability.

The WiSE-FT-LP method leverages a linear interpolation strategy to blend pre-trained and fine-tuned model weights, optimizing task-specific performance and overall model robustness. This approach involves carefully selecting the best interpolation coefficient and fine-tuning task-specific layers to achieve high-performance results while ensuring adaptability and stability across different practical scenarios and challenges.

\begin{table*}
  \caption{Robust Fine-tuning Evaluation for Downstream Point Cloud Classification on the ScanObjectNN Dataset: Values at the maximum (and tied maximum) positions are bolded, and values at the second maximum (and tied second maximum) positions are underlined.}

  \label{tab:rfte}
  \begin{tabular}{c|c|ccc|c}
    \toprule’
    \multirow{2}{*}{\textbf{Fine-tuning Method}}
    & \multirow{2}{*}{\textbf{Pre-training Method}}
    & \multicolumn{3}{c|}{\textbf{Overall Accuracy (\%)}}
    & \multirow{2}{*}{\textbf{Backbone Robustness (\%)}}\\

    && PB\_T50\_RS
    & OBJ\_ONLY
    & OBJ\_BG\\
    
    \midrule
    
    \multirow{2}{*}{Linear Probing}
    & Point-M2AE \cite{DBLP:conf/nips/ZhangG0FZW0022}
    & 77.86 & 86.06 & 84.85 & \textbf{92.9}\\
    & ReCon \cite{DBLP:conf/icml/QiDFGZMY23}
    & 83.80 & 90.71 & 90.62 & \textbf{92.34}\\
    \midrule
    
    \multirow{6}{*}{Full Fine-tuning}
    & \multirow{3}{*}{Point-M2AE \cite{DBLP:conf/nips/ZhangG0FZW0022}}
    & \underline{86.43} &&& 89.59\\
    &&& \underline{88.81} && 89.14\\
    &&&& \textbf{91.22} & 88.82\\
    & \multirow{3}{*}{ReCon \cite{DBLP:conf/icml/QiDFGZMY23}}
    & \textbf{91.26} &&& 91.82\\
    &&& \textbf{93.80} && 91.00\\
    &&&& \textbf{95.35} & 91.65\\
    
    \midrule
    
    \multirow{6}{*}{WiSE-FT \cite{DBLP:conf/cvpr/WortsmanIKLKRLH22} with greedy $\alpha$}
    & \multirow{3}{*}{Point-M2AE \cite{DBLP:conf/nips/ZhangG0FZW0022}}
    & 86.19 &&& 89.71\\
    &&& 87.95 && 89.75\\
    &&&& \textbf{91.22} & \underline{89.47}\\
    & \multirow{3}{*}{ReCon \cite{DBLP:conf/icml/QiDFGZMY23}}
    & \underline{90.74} &&& \underline{92.14}\\
    &&& 87.78 && \underline{91.13}\\
    &&&& 94.32 & 91.86\\
    
    \midrule
    
    \multirow{6}{*}{\textbf{WiSE-FT-LP (ours)} with greedy $\alpha$}
    & \multirow{3}{*}{Point-M2AE \cite{DBLP:conf/nips/ZhangG0FZW0022}}
    & \textbf{86.99} &&& \underline{90.68}\\
    &&& \textbf{89.33} && \underline{91.29}\\
    &&&& 90.02 & \underline{89.47}\\
    & \multirow{3}{*}{ReCon \cite{DBLP:conf/icml/QiDFGZMY23}}
    & \underline{90.74} &&& \underline{92.14}\\
    &&& \underline{92.60} && \underline{91.13}\\
    &&&& \underline{95.01} & \underline{91.94}\\
    
    \bottomrule
  \end{tabular}
\end{table*}

\subsection{Backbone Network Robustness Assessment}
The linear SVM evaluation method is widely employed in various applications \cite{DBLP:conf/colt/BoserGV92, DBLP:books/daglib/0097035}. This technique is utilized not only for assessing the discriminative capability of pre-trained features \cite{DBLP:conf/iccv/GoyalM0M19}, but also finds extensive usage across diverse fields. The fundamental principle involves evaluating feature effectiveness by constructing a linear classifier on training data to assess their performance in classification tasks. Delineating a hyperplane within the feature space to maximize separation between categories, SVM effectively classifies data.

In this study, linear SVM classification accuracy was used as an indicator for evaluating model robustness. This evaluation offers a more comprehensive measure of the model's performance, encompassing its stability against varying data distributions and noise. Such an approach aids in gaining insights into the model's real-world performance.

\section{Experiments}

\subsection{Experimental Setup}

In this paper, ModelNet \cite{DBLP:conf/cvpr/WuSKYZTX15} and ScanObjectNN \cite{DBLP:conf/iccv/UyPHNY19} are selected to evaluate the model robustness and downstream task performance. Among them, ModelNet is one of the classic datasets for synthetic 3D object recognition, containing about 12,000 meshed 3D CAD objects divided into 40 (ModelNet40) or 10 (ModelNet10) categories. This paper evaluates the model on the ModelNet40 data set, mainly using linear SVM classification learning and few-shot classification learning to evaluate the robustness of the model. ScanObjectNN is one of the most challenging 3D datasets, covering approximately 15,000 real-world objects from 15 categories. This paper uses this dataset as a downstream data distribution, targeting the classification task, and adopts three widely used settings: the hardest setting (PB\_T50\_RS), the no background setting (OBJ\_ONLY), and the background setting (OBJ\_BG).

In addition, this paper also uses Point-M2AE \cite{DBLP:conf/nips/ZhangG0FZW0022} and ReCon \cite{DBLP:conf/icml/QiDFGZMY23}'s pre-trained model parameters which are pre-trained on the point cloud data set ShapeNet \cite{DBLP:journals/corr/ChangFGHHLSSSSX15}, and their fine-tuned model parameters under three different settings on the ScanObjectNN \cite{DBLP:conf/iccv/UyPHNY19} dataset. During the fine-tuning process, this paper uses the widely used MLP-3 setting \cite{DBLP:conf/nips/ZhangG0FZW0022, DBLP:conf/icml/QiDFGZMY23, DBLP:conf/iclr/DongQZZSGYM23} to adjust the model head, that is, a Contains three consecutive layers of multi-layer perceptron. Such a setup has proven to be highly effective in similar tasks and better able to capture the complex features and patterns of the data.

\begin{figure*}[htb]
\centering
    \subfigure[Point-M2AE]{
        \includegraphics[width=\columnwidth]{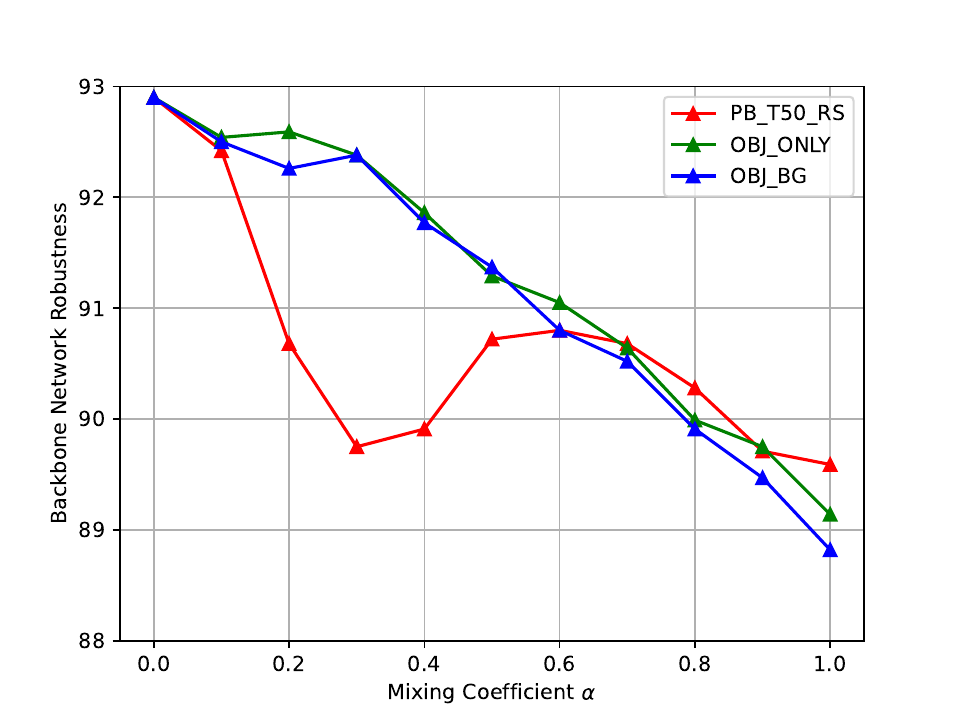}
        \label{pointm2ae_svm}
    }
    \subfigure[ReCon]{
        \includegraphics[width=\columnwidth]{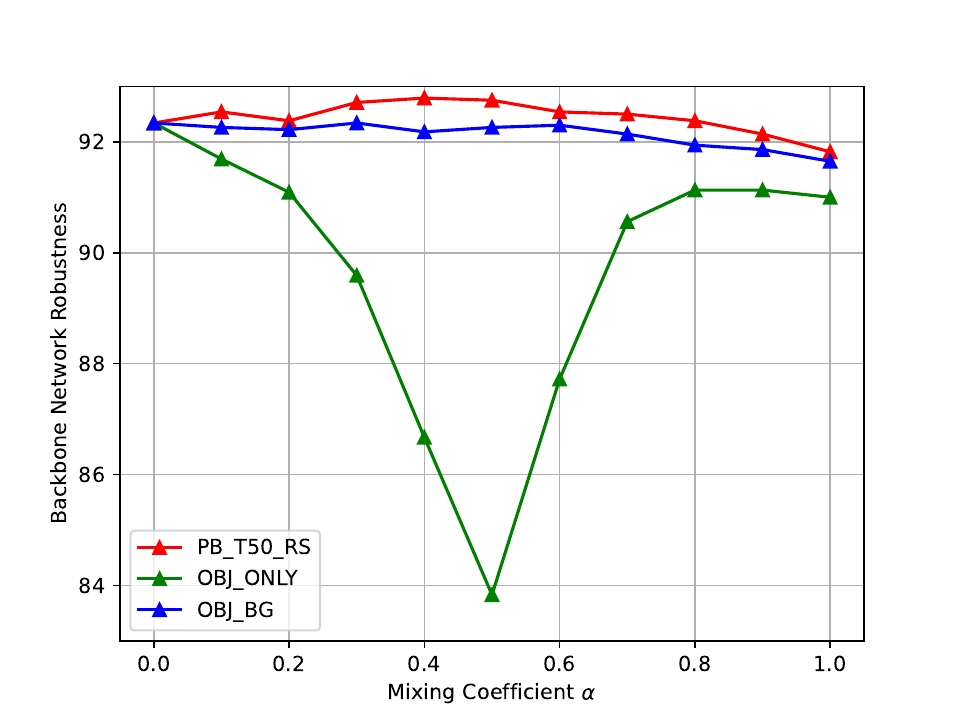}
        \label{recon_svm}
    }
\caption{Comparison of linear SVM classification results between Point-M2AE and ReCon on the ModelNet40 dataset: Three backbone networks fine-tuned on the ScanObjectNN dataset were compared through weight space integration. These networks correspond to the hardest setting (\textcolor{red}{red}), no background setting (\textcolor{green}{green}), and background setting (\textcolor{blue}{blue}).}
\label{svm}
\end{figure*}

\subsection{Robustness and Performance Evaluation}

This section presents a detailed analysis of quantitative performance and robustness evaluations for downstream point cloud classification tasks on the ScanObjectNN dataset, as illustrated in Table \ref{tab:rfte}. The table provides insights into the performance metrics of various fine-tuning methods and pre-training strategies across different task categories, along with their impact on the robustness of the backbone network.

We adopted a greedy mixing coefficient $\alpha$ selection strategy, starting from 1 and decreasing by 0.1 each time. If the current downstream task performance decreases by more than 0.1\%, or if model robustness begins to decline, we stop the decrementing process.

The analysis of Table \ref{tab:rfte} leads to the following conclusions:

\begin{itemize}
    \item Similar to some previous observations in the research \cite{DBLP:conf/iclr/KumarRJ0L22}, both Linear Probing and Full Fine-tuning fail to simultaneously ensure downstream task performance and backbone network robustness. Linear Probing prioritizes backbone network robustness, while Full Fine-tuning focuses solely on optimizing downstream task performance.
    
    \item Consistent with our intuition, WiSE-FT demonstrates better preservation of backbone network robustness compared to Full Fine-tuning. However, in most cases, WiSE-FT sacrifices downstream task performance.
    
    \item Comparatively, except for specific scenarios (Point-M2AE + OBJ\_BG), WiSE-FT-LP outperforms WiSE-FT in achieving better downstream task performance and backbone network robustness across most cases.
    
    \item In the majority of cases, WiSE-FT-LP strikes a balance by sacrificing less downstream task accuracy, achieving the best compromise between downstream task accuracy and backbone network robustness. In some scenarios, WiSE-FT-LP can simultaneously achieve superior downstream task accuracy and backbone network robustness, surpassing Full Fine-tuning.
\end{itemize}

\begin{figure*}[htb]
\centering
    \subfigure[The change curve of downstream task accuracy as the weight mixing coefficient $\alpha$ changes.]{
        \includegraphics[width=\textwidth]{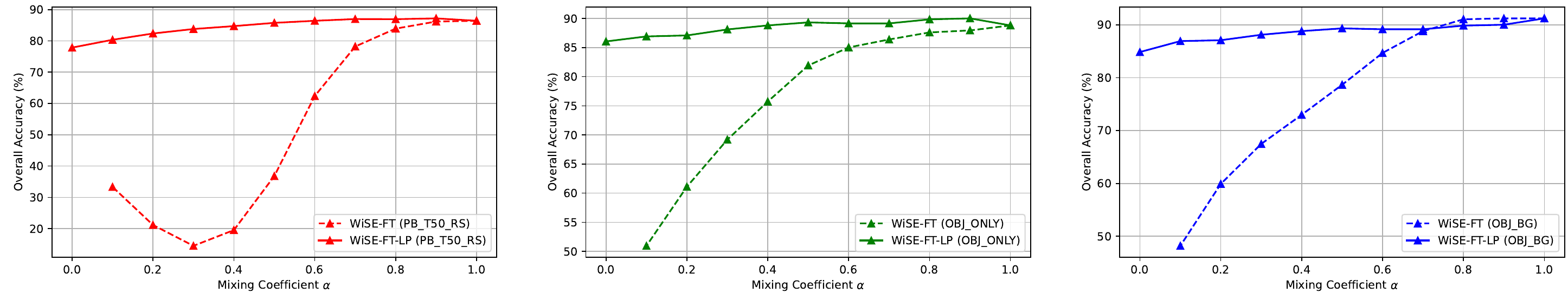}
        \label{pointm2ae_scan}
    }
    \subfigure[The change curve of backbone network linear SVM classification accuracy as the downstream accuracy changes. The superior coefficients are labeled in the figure, with coefficients of WiSE-FT-LP shown in regular font and coefficients of WiSE-FT shown in \textit{italics}.]{
        \includegraphics[width=\textwidth]{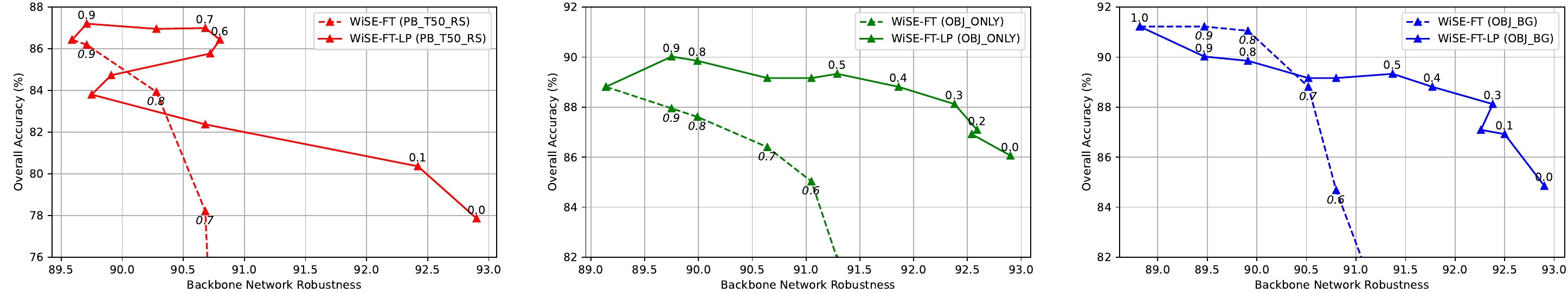}
        \label{pointm2ae_scan_svm}
    }
\caption{Comparison of the results of robust fine-tuning of Point-M2AE on three ScanObjectNN dataset settings. \textcolor{red}{Red}, \textcolor{green}{green}, and \textcolor{blue}{blue} represent the most difficult settings, settings without background, and settings with background respectively. The dashed line in the figure represents the original WiSE-FT, while the solid line represents the proposed WiSE-FT-LP.}
\label{pointm2ae_wise}
\end{figure*}

\subsection{Backbone Network Robustness Evaluation}

In the experiments, a series of intermediate models were generated by interpolating the pre-trained model and the fine-tuned model. The interpolation coefficient $\alpha$ is interpolated from 0 to 1 at an interval of 0.1. A total of 9 new interpolation models are obtained, plus the pre-training model and the full fine-tuning model, for a total of 11 models. Among these models, when $\alpha=0$ represents the pre-trained model, while $\alpha=1$ represents the full fine-tuning model. As shown in Figure \ref{svm}, through this experimental design, the performance of the model at different interpolation coefficients can be observed. These results are important for an in-depth understanding of the model's characteristics and for optimizing the model.

By analyzing Figure \ref{svm}, it can be observed that no matter what pre-training method and fine-tuning task selection is used, the general trend is that with the mixing coefficient $\alpha$ ($\alpha=1$ vs. $\alpha=0$) increases, the linear SVM classification accuracy of the backbone network also decreases, which means that the robustness of the model decreases. This illustrates the limitations of only full-volume fine-tuning. Furthermore, in Figure \ref{pointm2ae_svm}, in the hardest setting (PB\_T50\_RS), the classification accuracy reaches the lowest point at $\alpha=0.3$, while in Figure \ref{recon_svm}, no background setting (OBJ\_ONLY), the classification accuracy reaches its lowest point around $\alpha=0.5$. The sharp decrease in the accuracy of linear SVMs indicates that the quality of model features obtained through weight space interpolation has reached a nadir. These interpolation coefficients may indicate that the model has a conflict between robustness and downstream task performance, so interpolating around this value may make it difficult to obtain a backbone network that balances robustness with downstream task performance. Overall, except for individual minimum points, for the three downstream task setting models of Point-M2AE, smaller $\alpha$ often means stronger model robustness.

Further analysis of Figure \ref{recon_svm} indicates that in models with downstream fine-tuning of ReCon, using no background setting (OBJ\_ONLY) versus background setting (OBJ\_BG), smaller $\alpha$ values generally correspond to stronger model robustness. Specifically, the model fine-tuned at the hardest setting achieved maximum classification accuracy at $\alpha = 0.4$ (92.79\% versus 91.82\%), suggesting that models interpolated around this coefficient may exhibit better robustness. By interpolating around $\alpha = 0.4$, a fixed-parameter backbone network with improved robustness and downstream task performance can be obtained. However, it is worth noting that the linear SVM classification accuracy of the interpolation model without background setting (OBJ\_ONLY) reaches a locally low value at $0.2 \leq \alpha \leq 0.7$. This indicates that interpolation models within this range are affected by interpolation and are less robust. This phenomenon may be attributed to the conflict between the optimization directions of ReCon's pre-training tasks and the optimization directions of downstream tasks without background settings. Therefore, when selecting interpolation coefficients, it is essential to balance the robustness and performance of the model to achieve optimal fine-tuning results.

\subsection{Further Analysis for Point-M2AE}

In Figure \ref{pointm2ae_wise}, this paper explores the comparison of the results of robust fine-tuning of Point-M2AE on three ScanObjectNN dataset settings. Specifically, the following two curves are used to express the results: 
\begin{itemize}
    \item The curve where the downstream accuracy changes with the mixing coefficient $\alpha$.
    \item The curve where the backbone network linear SVM classification accuracy changes with the downstream accuracy.
\end{itemize}
These curves help understand the performance of the Point-M2AE model in different mixing coefficients under different downstream task data settings, and evaluate the robustness of its corresponding interpolation backbone network.

\subsubsection{Downstream Task Accuracy Change Curve}

As shown in Figure \ref{pointm2ae_scan}, as the mixing coefficient $\alpha$ decreases, the accuracy of the downstream tasks of the WiSE-FT fine-tuned model decreases significantly. Compared with the WiSE-FT fine-tuned model, the downstream task accuracy of the WiSE-FT-LP fine-tuned model is higher and is higher than or closer to the downstream task accuracy of the fully fine-tuned model when $\alpha=1$. In the hardest setting (PB\_T50\_RS), when $0.7\leq\alpha\leq 0.9$, the downstream task accuracy reaches 86.9\%, 86.95\%, 87.20\%, which is higher than the full fine-tuning model (when $\alpha=1$). When $0.5\leq\alpha\leq 0.9$, the downstream task accuracy reached 89.33\%, 89.16\%, 89.16\%, 89.85\%, and 90.02\% respectively, which is higher than the 88.81\% accuracy of the downstream tasks of the fully fine-tuned model (when $\alpha=1$); in the background setting (OBJ\_BG), when $\alpha=0.9$, The accuracy of downstream tasks reaches 91.22\%, which is equal to 91.22\% of the accuracy of downstream tasks of the full fine-tuning model (when $\alpha=1$).

It can be seen that through WiSE-FT-LP, the performance of downstream tasks including the initialization head model has been significantly improved. In addition, it was found that different downstream fine-tuning models have different adaptability to interpolation. Some models have a high space for improvement in downstream task performance, while the downstream task performance of some models' backbone networks has not been improved after weight space integration. This may be related to The differences in optimization directions between pre-training tasks and downstream tasks are related. In other words, some fine-tuned models may not get a "free lunch" from robust fine-tuning, and for these models, improvements in robustness conflict with downstream task accuracy. Therefore, how to balance robustness and downstream task accuracy has become a difficult problem faced by some models.

\begin{figure*}[htb]
\centering
    \subfigure[The change curve of downstream task accuracy as the weight mixing coefficient $\alpha$ changes.]{
        \includegraphics[width=\textwidth]{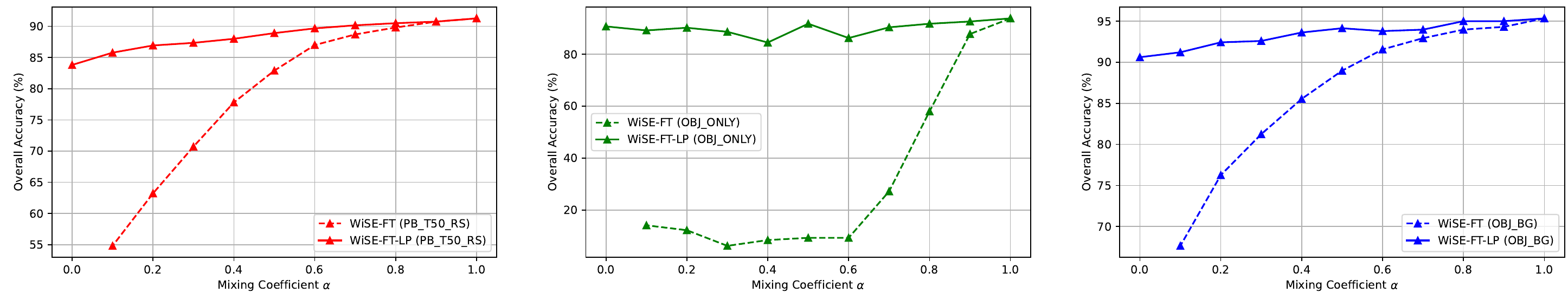}
        \label{recon_scan}
    }
    \subfigure[The change curve of backbone network linear SVM classification accuracy as the downstream accuracy changes. The superior coefficients are labeled in the figure, with coefficients of WiSE-FT-LP shown in regular font and coefficients of WiSE-FT shown in \textit{italics}.]{
        \includegraphics[width=\textwidth]{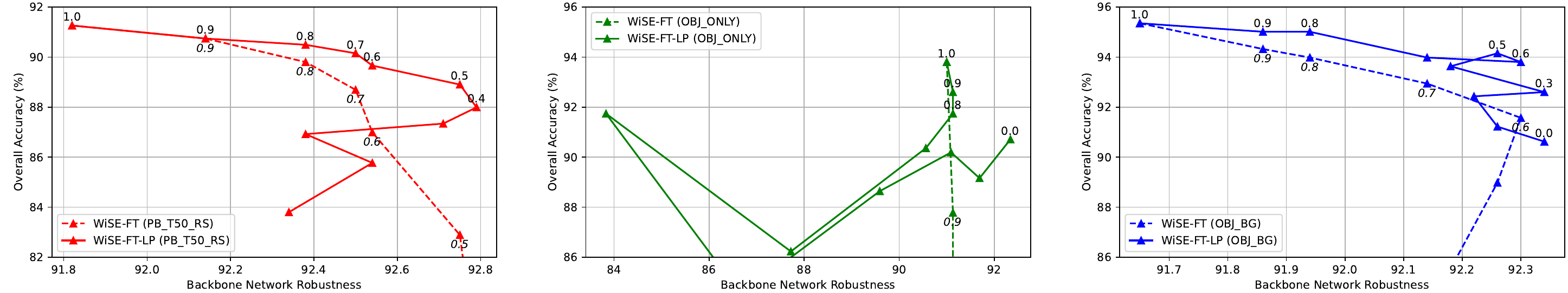}
        \label{recon_scan_svm}
    }
\caption{Comparison of the results of robust fine-tuning of ReCon on three ScanObjectNN dataset settings. \textcolor{red}{Red}, \textcolor{green}{green}, and \textcolor{blue}{blue} represent the most difficult settings, settings without background, and settings with background respectively. The dashed line in the figure represents the original WiSE-FT, while the solid line represents the proposed WiSE-FT-LP.}
\label{recon_wise}
\end{figure*}

\subsubsection{Mixing Coefficient Change Curve}

In Figure \ref{pointm2ae_scan_svm}, the curve depicting the linear SVM classification accuracy of the backbone network relative to downstream accuracy is presented. Firstly, it is evident that the interpolation curve formed by the WiSE-FT-LP fine-tuned model is positioned closer to the upper right corner compared to the WiSE-FT fine-tuned model, indicating that the proposed method achieves a superior balance between robustness and downstream task performance.

Secondly, despite the existence of interpolation curves, it is recognized that simultaneously enhancing the model's robustness and the performance of downstream tasks presents a significant challenge. In the hardest setting, with $\alpha = 0.7, 0.8, 0.9$, the downstream task accuracy and model robustness achieved $(86.90\%, 90.68\%)$, $(86.95\%, 90.28\%)$, $(87.20\%, 89.71\%)$, respectively, all surpassing the downstream task accuracy of $86.43\%$ and model robustness of $89.59\%$ achieved by the fully fine-tuned model ($\alpha = 1$).

In the setting without background, when $\alpha=0.5,0.8,0.9$, the downstream task accuracy and model robustness reached (89.33\%, 91.29\%), (89.85\%, 89.99\%), (90.02\%, 89.75\%), all higher than the downstream task accuracy of 88.81\% and model robustness of 89.59\% achieved by the fully fine-tuned model (when $\alpha=1$).

In the setting with background, when $\alpha=0.1, 0.3, 0.4, 0.5, 0.8, 0.9$, the downstream task accuracy and model robustness reached (86.92\%, 92.50\%), (88.12\%, 92.38\%), (88.81\%, 91.77\%), (89.33\%, 91.37\%), (89.85\%, 89.91\%), (90.02\%, 89.47\%), demonstrating improved downstream task accuracy compared to the fully fine-tuned model (when $\alpha=1$) which achieved an accuracy of 91.22\% and a model robustness of 88.82\%. 

Through the WiSE-FT-LP method, the model's robustness is significantly enhanced. Furthermore, it was observed that different downstream fine-tuning models exhibit varying adaptability to interpolation. Some models exhibit greater potential for improving robustness, while for others, the robustness conflicts with downstream tasks, likely attributed to differences between pre-training and downstream tasks and their optimization directions.

Experiments show that when $\alpha$ is closer to 1, the model has a higher probability of being able to simultaneously obtain better robustness and downstream task performance from WiSE-FT-LP, which is consistent with WiSE-FT \cite{DBLP:conf/cvpr/WortsmanIKLKRLH22}, when $\alpha$ is close to 0.5, the model can obtain better robustness to data offset. The results are different. This paper speculates that the reason for this phenomenon may be firstly due to the difference in fine-tuning between the head-based inference model and the zero-shot inference model. Secondly, it may also be caused by the difference in the adaptability of the image vision model and the point cloud vision model to the weight space integration. In the following experiments, this paper will replace the point cloud pre-training model and explore the performance of WiSE-FT-LP when $\alpha$ is close to 0.5.

\begin{figure*}[htb]
    \includegraphics[width=\textwidth]{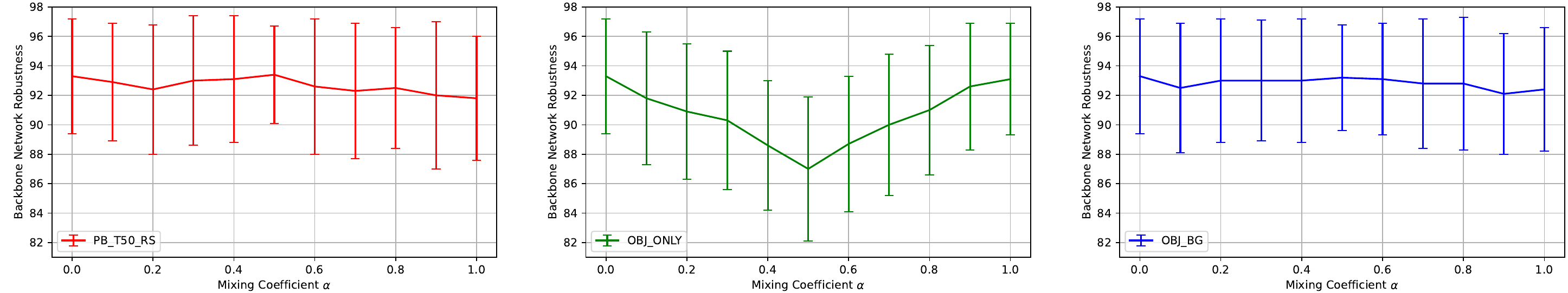}
    \caption{Comparison of ReCon's few-shot classification learning results on the ModelNet40 dataset: Three backbone networks fine-tuned on the ScanObjectNN dataset were compared through weight space integration. \textcolor{red}{Red}, \textcolor{green}{green}, and \textcolor{blue}{blue} represent the most difficult settings, settings without background, and settings with background respectively.}
    \label{recon_fsl}
\end{figure*}

\subsection{Further Analysis of ReCon}

In Figure \ref{recon_wise}, the study delves deeper into the robustness fine-tuning of ReCon across three different ScanObjectNN dataset settings. The comparison of results in this section is particularly noteworthy. It's essential to emphasize that specific parameter combinations were meticulously chosen during the experiment to ensure clarity and simplicity in interpreting the outcomes.

\subsubsection{Downstream Task Accuracy Change Curve}

In Figure \ref{recon_scan}, it is evident that as the weight mixing coefficient $\alpha$ decreases, the accuracy of the WiSE-FT fine-tuned model in downstream tasks experiences a notable decline, which aligns with previous experimental findings. In contrast, the accuracy of the WiSE-FT-LP fine-tuned model remains relatively stable and, in certain instances, even surpasses the downstream task accuracy of the fully fine-tuned model (when $\alpha=1$).

\subsubsection{Mixing Coefficient Change Curve}

Moreover, Figure \ref{recon_scan_svm} displays the curve depicting the linear SVM classification accuracy of the backbone network relative to downstream accuracy. The interpolation curve formed by the WiSE-FT-LP fine-tuning model approaches the upper right corner, indicating superior robustness and more balanced performance in downstream tasks. This observation provides valuable insights into the fine-tuning process, where model performance can be adjusted by modulating the interpolation coefficient $\alpha$.

The WiSE-FT-LP fine-tuning method offers advantages over the traditional WiSE-FT approach by achieving a better balance between model robustness and downstream task performance, ultimately enhancing model robustness while maintaining satisfactory performance levels.

In the hardest setting, when $\alpha=0.4$ and $\alpha=0.5$, the downstream task accuracy reaches 87.99\% and 88.90\% respectively, compared to 91.26\% for the fully fine-tuned model (when $\alpha=1$). Despite a slight decrease in accuracy, the model's robustness improves to 92.79\% and 92.75\% respectively, compared to 91.82\%, achieving a better balance. Similarly, in the setting with background, when $\alpha=0.3$ and $\alpha=0.5$, the downstream task accuracy achieves 92.60\% and 94.15\% respectively, compared to 95.35\% for the fully fine-tuned model (when $\alpha=1$). Despite a slight decrease in accuracy, the model's robustness improves from 91.65\% to 92.34\% and 92.26\%, maintaining a good balance.

Based on these results, WiSE-FT-LP reaffirms that in most cases, model robustness can be enhanced by sacrificing a small amount of downstream task accuracy. However, in the setting without background, when $\alpha=0.5$, both the downstream task accuracy and model robustness experience comprehensive degradation, specifically (91.74\%, 83.83\%), indicating that linear interpolation fails to improve model robustness. Therefore, it is necessary to discard the fine-tuned model and instead perform linear probe fine-tuning ($\alpha=0$) on the pre-trained model to ensure backbone network robustness. Additionally, these experimental findings suggest that coefficients near $\alpha=0.5$ may not be suitable for WiSE-FT-LP weight space interpolation, with $\alpha$ values closer to 0.9 often achieving a better balance between downstream task accuracy and model robustness.

\subsection{Alternative Robustness Assessment}

In addition to employing linear SVM evaluation on different data distributions, such as ModelNet40, to assess the model's robustness, this paper also explores using few-shot classification results to characterize the backbone network's robustness. As illustrated in Figure \ref{recon_fsl}, the interpolation curves across the three ScanObjectNN settings align with the findings depicted in Figure \ref{svm}, indicating that the outcomes of few-shot learning serve as an indicator of the backbone network's robustness.

However, it's important to note that few-shot inference demands more computational resources and time than linear SVM classification. Therefore, for efficiency considerations, this paper opts for utilizing the latter method to evaluate the robustness of the model's backbone network.

\section{Conclusion}
Our study introduces a robust fine-tuning method for pre-trained 3D point cloud models, tailored to improve feature robustness in downstream tasks. We address inherent limitations of existing fine-tuning methods, particularly in learning robust models amidst diverse challenges. Our proposed approach, Weight-Space Ensembles for Fine-Tuning then Linear Probing (WiSE-FT-LP), integrates the original pre-training and fine-tuning models through weight-space integration followed by Linear Probing. This innovative method significantly improves the performance of downstream fine-tuned models under distribution shifts, while maintaining high performance on the target distribution. Additionally, we leverage insights from few-shot learning to inform our approach.

We apply the WiSE-FT-LP method to mainstream 3D point cloud pre-trained models, conducting comprehensive evaluations to assess model parameter quality and downstream task performance degradation. The experimental results vividly demonstrate the effectiveness of WiSE-FT-LP in enhancing model robustness. Our method successfully balances downstream task performance and model feature robustness without necessitating changes to the underlying model structures. These findings underscore the importance of robust fine-tuning methods for advancing the reliability and effectiveness of pre-trained 3D point cloud models in diverse applications.


\bibliographystyle{ACM-Reference-Format}
\bibliography{sample-base}










\end{document}